\documentclass[11pt, letterpaper]{berkeley}
\usepackage{parskip}

\makeatletter
\def\mathcolor#1#{\@mathcolor{#1}}
\def\@mathcolor#1#2#3{%
  \protect\leavevmode
  \begingroup
    \color#1{#2}#3%
  \endgroup
}
\makeatother

\usepackage{graphicx}
\usepackage{amsmath,amssymb,amsfonts}
\usepackage{graphicx}
\usepackage{textcomp}
\usepackage[font=small]{caption}
\usepackage{float}
\usepackage{placeins}

\usepackage{multirow}
\usepackage{acronym}
\usepackage{subcaption}
\usepackage{xfrac}
\usepackage{latexcolors}
\usepackage{xspace}
\usepackage{wrapfig}

\usepackage[hyperpageref]{backref}
\def\BibTeX{{\rm B\kern-.05em{\sc i\kern-.025em b}\kern-.08em
    T\kern-.1667em\lower.7ex\hbox{E}\kern-.125emX}}

\usepackage{booktabs}
\usepackage{graphicx}
\usepackage{multirow}
\usepackage{arydshln}

\newcommand{\bz}{\mathbf{z}}

\newcommand{\bs}{\mathbf{s}}
\newcommand{\ba}{\mathbf{a}}

\newcommand{\bg}{\mathbf{g}}

\newcommand{\methodname}{PTR\xspace}

\newcommand{\systemname}{V-PTR}

\usepackage{amsmath,amsfonts,bm}

\def\eqref#1{equation~\ref{#1}}

\def\1{\bm{1}}

\DeclareMathAlphabet{\mathsfit}{\encodingdefault}{\sfdefault}{m}{sl}
\SetMathAlphabet{\mathsfit}{bold}{\encodingdefault}{\sfdefault}{bx}{n}

\newcommand{\E}{\mathbb{E}}

\newcommand{\tablestyle}[2]{\setlength{\tabcolsep}{#1}\renewcommand{\arraystretch}{#2}\centering\footnotesize}
\newlength\savewidth\newcommand\shline{\noalign{\global\savewidth\arrayrulewidth
  \global\arrayrulewidth 1pt}\hline\noalign{\global\arrayrulewidth\savewidth}}

\usepackage{xspace}

\usepackage{tabularx}

\title{Robotic Offline RL from Internet Videos \\via Value-Function Pre-Training}
\runningtitle{Robotic Offline RL from Internet Videos via Value-Function Pre-Training}

\reportnumber{} %

\author[*,1]{Chethan Bhateja}
\author[*,1]{Derek Guo}
\author[*,1]{Dibya Ghosh}
\author[1,2]{Anikait Singh}
\author[1]{Manan Tomar}
\author[2]{Quan Vuong}
\author[2]{Yevgen Chebotar}
\author[1]{Sergey Levine}
\author[1]{Aviral Kumar}

\affil[*]{Equal contributions}
\affil[1]{UC Berkeley}
\affil[2]{Google DeepMind}

\correspondingauthor{Chethan Bhateja (\href{mailto:chetbhateja@berkeley.edu}{chetbhateja@berkeley.edu}), Derek Guo  (\href{mailto:derekguo@berkeley.edu}{derekguo@berkeley.edu}), Dibya Ghosh (\href{mailto:dibya@berkeley.edu}{dibya@berkeley.edu})}

\begin{document}

\begin{abstract}
Pre-training on Internet data has proven to be a key ingredient for broad generalization in many modern ML systems. What would it take to enable such capabilities in robotic reinforcement learning (RL)? Offline RL methods, which learn from datasets of robot experience, offer one way to leverage prior data into the robotic learning pipeline. However, these methods have a ``type mismatch'' with video data (such as Ego4D), the largest prior datasets available for robotics, since video offers observation-only experience without the action or reward annotations needed for RL methods. In this paper, we develop a system for leveraging large-scale human video datasets in robotic offline RL, based entirely on learning value functions via temporal-difference learning. We show that value learning on video datasets learns representations that are more conducive to downstream robotic offline RL than other approaches for learning from video data. Our system, called V-PTR, combines the benefits of pre-training on video data with robotic offline RL approaches that train on diverse robot data, resulting in value functions and policies for manipulation tasks that perform better, act robustly, and generalize broadly. On several manipulation tasks on a real WidowX robot, our framework produces policies that greatly improve over prior methods. Our video and additional details can be found on our \href{https://dibyaghosh.com/vptr/}{project website}.
\end{abstract}

\maketitle

\begin{center}
    \vspace{.5cm}
    \includegraphics[width=\linewidth]{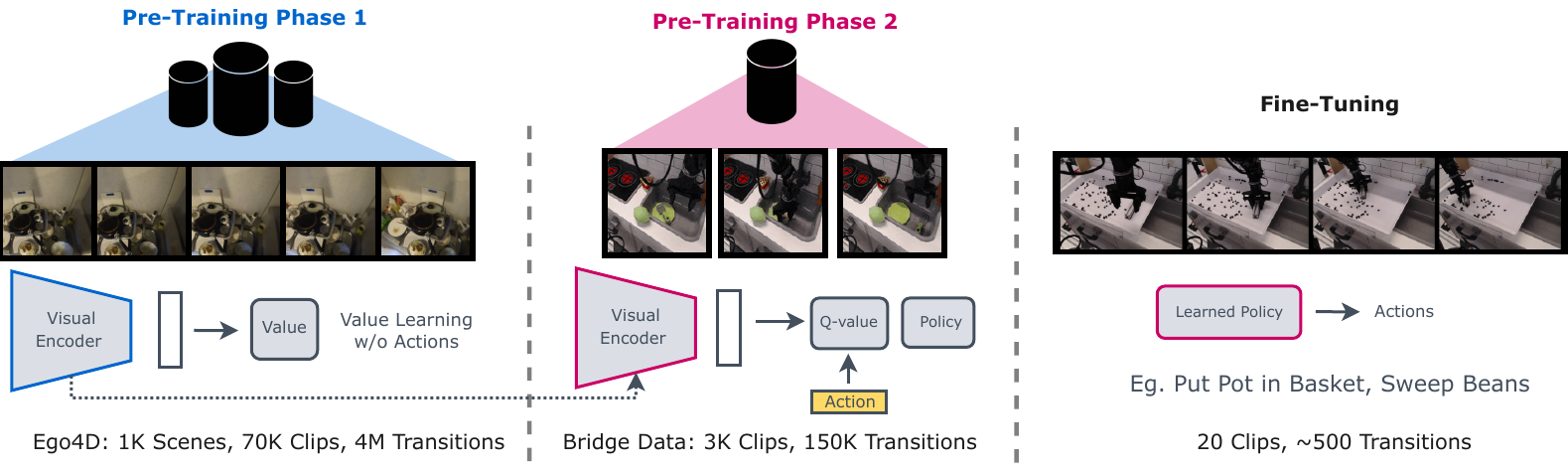}
           \captionof{figure}{\textbf{Video pre-training for robots (V-PTR)} pre-trains by learning value functions on a large-scale video dataset like Ego4D, and continues refining learned visual features by performing offline RL on multi-task robot data like the Bridge dataset. This pre-training provides a useful initialization for offline RL fine-tuning on downstream robot tasks with improved generalization and robustness compared to other approaches. }
    \label{fig:main_fig}
    \vspace{.5cm}
\end{center}

\vspace{-0.1cm}
\section{Introduction}
\vspace{-0.1cm}

Developing methods capable of acquiring robotic skills that generalize widely to new scenarios is an important problem in robotic learning. In other areas of machine learning, broad generalization has been fueled primarily by pre-training on large datasets with a diversity of behavior. It seems compelling that the same formula may be applied to robotic learning, but in practice, even our largest robotic datasets contain data for relatively few tasks, and from a limited number of scenarios. In principle, robotic reinforcement learning (RL) should be able to learn from more general sources of data like human video, which are far more abundant and capture a broader set of skills, situations, and interactions. However, these datasets are difficult to incorporate into RL methods that exist today, since internet-scale video data does not come with action or reward annotations present in typical robot data.  

Existing works~\citep{nair2022r3m,xiao2022masked,radosavovic2022real} include video data in the robot learning pipeline by performing self-supervised visual representation learning on video data ~\citep{he2111masked}, followed by downstream policy learning via behavioral cloning using the learned representations. While such an approach can extract visual features from video, it is limited in its ability to extract a deeper ``functional'' understanding of the world from video data: in principle, despite differences in embodiment, human videos can still be used to understand intents and affordances that can be executed in the real world, the dynamics, and the eventual outcomes that can be attained by acting. 

Motivated by the above desiderata for video pre-training, in this work, we aim to develop an approach that pre-trains on Internet-scale human video to produce representations for downstream offline RL. Our main contribution is a system, which we call \textbf{V}ideo \textbf{P}re-\textbf{T}raining for \textbf{R}obots (\systemname), that fits value functions to model long-term outcomes achieved when solving tasks on action-free video data. 

Concretely, \systemname\ pre-trains on human videos by learning an intent-conditioned value function \citep{ghosh2023reinforcement} via temporal-difference learning (TD-learning). This approach eschews self-supervised representation learning objectives utilized in prior works~\citep{nair2022r3m,ma2022vip,radosavovic2022real} in favor of a TD value learning objective, just as how downstream offline RL agents will fine-tune task-specific value functions. Next, we fine-tune on a multi-task robotic dataset, which is annotated with actions, tasks, and rewards, using value-based offline RL~\citep{kumar2022pre}. Downstream, when a target task is specified, \systemname\ fine-tunes the multi-task policy on this task. Each phase of our system gradually incorporates the knowledge of ``what future outcomes can be achieved'' (video pre-training), ``what robot actions lead to these outcomes'' (robot pre-training), and ``how can the desired task be solved'' (fine-tuning).

Our experiments on several manipulation tasks on a real WidowX robot show that by pre-training on human video data (Ego4D~\citep{grauman2022ego4d}) and multi-task robot data (Bridge data~\citep{ebert2021bridge}), \systemname\ endows downstream offline RL methods with significantly improved zero-shot generalization and robustness to different target objects, distractors, and other variations in the workspace compared to prior methods that learn from videos, significantly outperforming prior methods including VIP~\citep{ma2022vip}. To our knowledge, our work presents the first large-scale demonstration showing that TD-learning alone can be an effective approach to pre-train from video for robotic RL.

\vspace{-0.1cm}
\section{Related Work}
\label{sec:related}
\vspace{-0.1cm}

A number of prior approaches learn representations from video by applying image-level representation objectives on individual frames in the video or by modelling temporal dependencies along the frames in a given trajectory. The former includes objectives like reconstruction \citep{nair19ccrig, Seo2022MaskedWM, Xiao2022MaskedVP,karamcheti2023language} or contrastive learning on images \citep{Srinivas2020CURLCU}. While these objectives are widely used in computer vision, resulting representations do not capture any information about environment dynamics. The latter approaches model long-term dynamics from video by predicting the next frame \citep{Seo2022ReinforcementLW}, learning value functions \citep{ma2022vip,ghosh2023reinforcement}, or running time-contrastive learning~\citep{Sermanet2017TimeContrastiveNS,Nair2022R3MAU}. 
 
In the context of robot learning, recent works learn representations from internet video datasets like Ego4D~\citep{grauman2022ego4d}, using masked auto-encoders~\citep{nair2022r3m,karamcheti2023language}, language-video alignment~\citep{karamcheti2023language}, or time-contrastive learning~\citep{nair2022r3m, ma2022vip}, and train downstream policies on frozen features using behavioral cloning. In our experiments, we compare to several of these methods, and find that \systemname\ attains a higher performance on real-world tasks, especially when evaluated with high initial state variability and in the presence of distractors. Specifically restricted to the setting with egocentric human video, some recent work~\citep{bahl2023affordances} also attempts to utilize wrist trajectory prediction to determine intermediate waypoints to guide robot controllers in accomplishing a task. This approach is orthogonal to our goal extracting a state representation and our approach can, in principle, be utilized in conjunction with the method from this prior work.  

The most closely related work is value-implicit pre-training (VIP)~\citep{ma2022vip}, which pre-trains a value function using time-contrastive prediction for downstream reward shaping. Both learn value functions during pre-training, albeit with entirely different algorithms (contrastive learning vs. TD learning), for different policies (dataset policy vs. intent-conditioned policy), exhibiting different generalization properties~\citep{Baird1995, dann2014policy}. Furthermore, the system desiderata differ for VIP and \systemname: VIP focuses on learning good visual reward functions for weighted behavioral cloning, while we seek good value function initializations for downstream offline RL. In our experiments, we find that when both pre-training approaches are evaluated on the quality of downstream offline RL, those initialized with \systemname~improve over VIP in terms of generalization and robustness. While obtaining reward functions is not the focus of this paper, our experiments show that the V-PTR representation alone outperforms VIP with reward shaping.

Finally, a class of methods attempt to modify the downstream RL algorithm to train on video and robot data together in lieu of a video pre-training stage. For instance, \citep{Torabi2018BehavioralCF, Schmeckpeper2020ReinforcementLW, Baker2022VideoP, Chang2022LearningVF} train an inverse dynamics model to label video transitions with action pseudo-labels to use alongside the robotic experience; \citep{Stadie2017ThirdPersonIL, Torabi2018GenerativeAI} use inverse RL to imitate the state-distribution present in the video. These methods succeed only when a small domain gap exists between video data and robot data, so that observations from video data can plausibly be interpreted as robot data. This condition fails in our setup, as we utilize Ego4D~\citep{grauman2022ego4d} and the Bridge~\citep{ebert2021bridge} datasets, where observations in these datasets differ significantly from each other, including a major difference in viewpoint (e.g., egocentric view in video data~\citep{grauman2022ego4d} \& shoulder view in robot data~\citep{ebert2021bridge}). To our knowledge, no method of this type has been successfully applied to this setting.

\vspace{-0.1cm}
\section{Problem Statement and Background}
\label{sec:problem}
\vspace{-0.1cm}

We aim to leverage Internet-scale video data and multi-task robotic data to boost the robustness and generalization of robotic offline RL. We formulate the robot skill learning problem as the problem of maximizing infinite-horizon discounted reward in a Markov decision process (MDP).

\textbf{Formal problem statement.} We assume access to two pre-training datasets: an Internet-scale video dataset $\mathcal{D}_\text{video}$ (e.g., the Ego4D dataset~\citep{grauman2022ego4d}) and a target dataset, $\mathcal{D}_\text{target}$ of a limited number of demonstrations for a given target task on the robot. Additionally we are also provided a dataset of multi-task robot behaviors, $\mathcal{D}_\text{robot}$, which may not contain any data relevant to the target task. 
The video dataset $\mathcal{D}_\text{video}$ consists of sequences of frames (i.e., observations in the MDP), with no action or rewards. Denoting a frame as $\bs_{i, j}$, we define $\mathcal{D}_\text{video} := \left \{ \left(\bs_{i, 0}, \bs_{i, 1}, \cdots \right) \right\}_{i=1}^{n_\text{video}}$. The target dataset, $\mathcal{D}_{\text{target}}$, comprises of a few demonstrations of the target task on the robot $\mathcal{D}_\text{target} := \left \{ \left(\bs_{i, 0}, \ba_{i, 0}, r_{i, 0}, \bs_{i, 1}, \cdots \right)\right\}_{i=1}^{n_\text{target}}$, where the reward, $r_{i, j}$ is annotated to be +1 only on the final three timesteps of the demonstration (following \cite{kumar2022pre}). 
The multi-task robot dataset $\mathcal{D}_\text{robot}$ is organized identically to the target robot dataset, but with an additional task annotation on each trajectory $t_i$, which is specified either as a one-hot identifier or by natural language. Our goal is train policy $\pi$ which maximizes the $\gamma$-discounted cumulative reward, $\mathbb{E}_{\bs_0 \sim \rho_0, \ba_{0:\infty}, \bs_{1:\infty} \sim \pi} \left[ \sum_{t=0}^\infty \gamma^t r(\bs_t, \ba_t) \right]$, starting from a more diverse set of initial states indicated by the distribution $\rho_0$ than what was observed in the target dataset (e.g., more variation in distractor objects).

\textbf{Background.} Our system utilizes a generalized formulation of goal-conditioned RL and temporal-difference learning for pre-training value functions. In a nutshell, the goal-conditioned RL problem trains the agent to achieve arbitrary goal frames $\bg$, where rewards are specified by the sparse signal of $\mathbb{I}\left(\bs = \bg\right)$ when the frame is identical to the goal frame. Although the reward signal is sparse, goals and rewards can be defined by hindsight relabelling~\citep{andrychowicz2017hindsight}.
To learn a policy for the downstream task, we use value-based offline RL methods, which optimize $\pi$ against a learned Q-function $Q^\pi(\bs, \ba)$. The Q-value function measures the expected long-term reward attained when executing action $\ba$ at state $\bs$, then following policy $\pi$ thereafter, and satisfies the Bellman equation $Q^\pi(\bs, \ba) = r(\bs, \ba) + \gamma \mathbb{E}_{\bs', \ba'}[Q^\pi(\bs', \ba')]$. 

\section{Video Pre-Training for Robotic Offline RL}
\label{sec:videos}

Even though video data contains rich functional and dynamic information useful for downstream skill learning, this data cannot directly be integrated into robotic offline RL pipelines, which expect data annotated with the agent's actions and the reward signal. 
{In this section, we develop \systemname, our system that pre-trains general value functions on Internet-scale video data, with the intention of fine-tuning value functions for the desired downstream robotic task using value-based offline RL.} We first present an overview of our system next, and then discuss each of the components in detail subsequently. 

\begin{figure}
  \begin{center}    
  \includegraphics[width=0.5\linewidth]{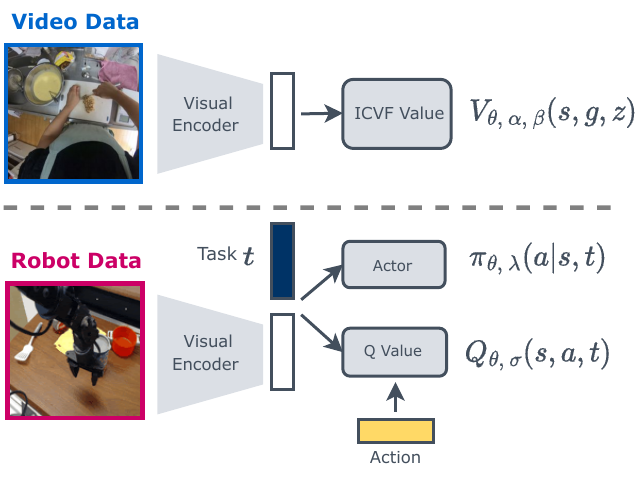}
  \end{center}
  \vspace{-0.4cm}
\caption{\footnotesize{\textbf{Network architecture}. \systemname\ first pre-trains image representations by training a general value function from video and then refines this representation via multi-task pre-training on robot data.}}
\vspace{-0.3cm}
\end{figure}
 \textbf{System overview.} Our system, \systemname, pre-trains in two phases: first on video data, and then on multi-task robot data. In the first phase, we train an intent-conditioned value function \citep{ghosh2023reinforcement} on action-less video data using a value-learning objective to model the outcomes associated with solving the parametric family of goal-achieving tasks. Next, we refine this representation with multi-task robot data with actions and rewards, by training a state-action Q-function on this representation using offline RL. We expect the video and multi-task robot data to capture dynamic features in the environment, with the robot data bridging the domain gap between human video and the robot, and aligning the learned representation with the agent's action space. Downstream, to adapt the system to a new target task, in our final phase, our system fine-tunes the Q-function and the policy on the target dataset.

\vspace{-0.05cm}
\subsection{Phase 1: Video Pre-Training via TD-Learning}

Since the goal of video pre-training is to improve the performance of downstream value-based offline RL, we turn to learning value functions on the video data as a natural pre-training procedure.
We choose to pre-train by learning an intent-conditioned value function (ICVF), a recently-proposed general value function that can be efficiently trained on passive data without action labels \citep{ghosh2023reinforcement}. An ICVF, annotated $V(\bs_\text{video}, \bg_\text{video}, \bz)$ computes the value obtained towards reaching a goal $\bg_\text{video}$, assuming the policy \emph{intended} to reach a different intended goal $\bz$, and is formally defined as 
\begin{equation*}
V(\bs_\text{video}, \bg_\text{video}, \bz) = \E_{a_t \sim \pi_\bz^*(\cdot | \bs_t)}\big[\sum_{t}\gamma^t \mathbb{I}\left(\bs_\text{video} = \bg_\text{video} \right) \big].
\end{equation*}
As with a standard value function, the ICVF can be learned by temporal-difference (TD) learning on its corresponding Bellman equation, using a target network to bootstrap the predictions of the learned value function. Specifically, for a given tuple $(\bs, \bg, \bz) \sim \mathcal{D}_\text{video}$, this TD objective function is given by:
\begin{align}
\label{eqn:icvf_objective_training}
    \min~ \Big[ \left( \alpha - \mathbb{I}\left(A \leq 0 \right) \right)\cdot \left( \mathbb{I}\left(\bs_\text{video} = \bg_\text{video} \right) + \gamma \bar{V}(\bs'_\text{video}, \bg_\text{video}, \bz) - V(\bs_\text{video}, \bg_\text{video}, \bz) \right)^2 \Big], \nonumber
\end{align}
where $A = \mathbb{I}\left(\bs_\text{video} = \bg_\text{video} \right) + \gamma \bar{V}(\bs'_\text{video}, \bz, \bz) - V(\bs_\text{video}, \bz, \bz)$ is the implied advantage of $\bs \to \bs'$ when acting to reach the observation $\bz$ and $\bar{V}$ is a delayed  copy of the value function (i.e., a target network). We follow the goal-sampling strategy from ICVF: after sampling a start frame $\bs$ from the video dataset, we choose the goal $\bg$ to be either the next observation, a future observation in the video, or a random observation from a different video. The intent $\bz$ is also an observation appearing in the video, and is chosen in a similar fashion as the goal state. Additionally, following \citet{ghosh2023reinforcement}, with some probability we set $\bz = \bg$. 

We follow \citet{ghosh2023reinforcement} and parameterize our estimated value function as
\begin{equation*}
    V(\bs_\text{video}, \bg_\text{video}, \bz) := \phi(\bs_\text{video})^\top T(\bz) \psi(\bg_\text{video}),
    \label{eqn:icvf_structure}
\end{equation*}
where $\phi_\theta$ and $\psi_\alpha$ denote models that transform the observation and the goal observation respectively into low-dimensional representations, and $T_\beta$, a learned mapping aligning the two representations. At convergence, the ICVF provides a measure of temporal spatiality, and the learned representation $\phi_\theta(\bs)$ offers a useful feature basis for downstream value functions.

\subsection{Phase 2: Multi-Task Robot Pre-Training via Offline RL}

In the next phase, we refine the learned representation on a multi-task robot dataset, $\mathcal{D}_\text{robot}$, to narrow the domain gap between robot image observations and human video, and to provide information about the target robot embodiment (i.e., the actions affordable by the robot). The purpose of this phase is to provide information about the target robot embodiment: crucially note that the tasks and workspaces in this robot dataset are explicitly disjoint from the target tasks used in the downstream evaluation.

\systemname\ uses multi-task robot data to pre-train a Q-function and a policy using multi-task conservative Q-learning (CQL)~\citep{kumar2022pre}, initializing the parameters of \emph{both} the Q-function and the policy using the backbone learned during video pre-training in Phase 1. Concretely, the Q-function and the policy are conditioned on the pre-trained representation of the robot observation $\phi_\theta(\bs_\text{robot})$ alongside a task vector $t$ (either a one-hot task identifier or a language embedding from a sentence transformer). At the onset of this phase, we initialize the representation encoder $\phi_\theta$ to the encoder $\phi_{\theta^*_\text{video}}$ obtained at the end of phase 1.
The value function is trained to satisfy the Bellman equation, 
\begin{align*}
\label{eqn:cql_training}
\min_{\theta}~ & \alpha \cdot \mathcal{L}_\text{CQL}(\theta) + \mathbb{E}_{\mathcal{D}_\text{robot}}\left[\left(Q_{\theta}(\bs, \ba; t) - r - \gamma \bar{Q}(\bs', \ba', t)\right)^2 \right],
\end{align*}
with target Q-network $\bar{Q}$, and CQL regularizer $\mathcal{L}_\text{CQL}(\theta)$ ~\citep{kumar2020conservative}, and the policy trained to maximize value, 
\begin{align*}
    \max_\lambda~~~~ \mathbb{E}_{\mathcal{D}_\text{robot}}\left[ \mathbb{E}_{\ba \sim \pi_{\lambda}(\cdot|\bs; t)}[Q(\bs, \ba; t)]  \right] + \beta \mathcal{H}(\pi_{\lambda}).
\end{align*}
After pre-training, we have a multi-task Q-function and policy that can be fine-tuned to the desired downstream task using a small target dataset.

\subsection{Phase 3: Fine-Tuning to a Target Task}

Finally, we fine-tune the value function and policy from the pre-training stage to the target task by running CQL \citep{kumar2020conservative} on the target dataset $\mathcal{D}_\text{target}$. We follow \citet{kumar2022pre} and treat the target data simply as a new task; fine-tuning involves assigning a new task identifier to the target data (either a new one-hot vector or a new language command), and continuing to run multi-task offline CQL on the robot pre-training and target tasks jointly. To address any overfitting induced by the small size of $\mathcal{D}_\text{target}$, we perform stratified sampling between the multi-task robot data $\mathcal{D}_\text{robot}$ and  the target data $\mathcal{D}_\text{target}$: $1 - \tau$ proportion of the training batch comes from $\mathcal{D}_\text{robot}$ and $\tau$ from $\mathcal{D}_\text{target}$, where $\tau$ is small (we use $\tau = 0.1$). 

\subsection{Implementation Details}

\noindent \textbf{\textit{Video pre-training:}} We use video frames at a $224\!\!\times\!\!224$ resolution. The three components parameterizing the ICVF are implemented as separate $3$-layer MLP heads on a shared visual backbone encoder. We use a Resnetv2-50 \citep{He2016IdentityMI} as our backbone since smaller convolutional networks led to worse pre-training performance. We replace all batch normalization layers with group normalization~\citep{wu2018group}, since prior works~\citep{bhatt2019crossnorm, kumar2022pre} often found batch normalization to be unstable. To avoid overfitting to spurious correlations between consecutive frames in a video clip, we use image augmentation (random crops and color jitter)~\citep{kostrikov2020image}, weight decay, and dropout. We train the model for $2\times10^6$ gradient steps with a batch size of $64$, using Adam, and learning rate $10^{-4}$ cosine annealed through training. All remaining designs we take from the open-source code~\citep{ghosh2023reinforcement}.
\vspace{-0.1em}

\noindent \textbf{\textit{Multi-task robot pre-training:}} We fine-tune on the multi-task robot dataset primarily following design decisions from \citet{kumar2022pre}. The CQL policy takes the RGB image, proprioceptive state, and task identifier as input. The RGB image is passed through a visual encoder, then concatenated with the remaining information, and passed through a $2$-layer MLP. We additionally concatenate task and action information into each hidden layer of the MLP. The CQL value function is parameterized similarly, but it also takes the action vector as input. Encoder parameters for both the value and policy are initialized from the video pre-trained ICVF, but there is no further weight tying between the networks. We train multi-task CQL for $2 \times 10^5$ gradient steps with batch size of $64$, and Adam with a constant learning rate of $10^{-4}$.

\section{Experimental Results}
\label{sec:result}

The goal of our experiments is to validate the effectiveness of \systemname\ in boosting the generalization and robustness of robotic offline RL. We evaluate \systemname\ in several scenarios requiring generalization to new scenes, compare to other approaches for incorporating video data, and perform additional diagnostic experiments to understand how value pre-training can provide useful representations for downstream robotic RL. Our settings include several challenging robotic manipulation tasks that require zero-shot generalization to new objects and distractors. A video of our evaluations and diagnostic experiments can be found on our \href{https://sites.google.com/view/v-ptr}{\textbf{project website}}.

\begin{figure}
    \centering
    \vspace{0.1cm}
    \includegraphics[width=0.9\linewidth]{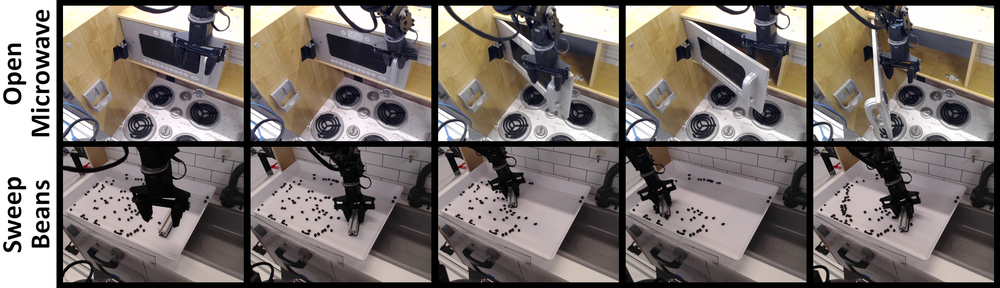}
    \vspace{-0.1cm}
    \caption{\label{fig:complex_tasks} \footnotesize{\textbf{Examples of setup and successful rollouts for complex tasks}. We utilize the robot setup from the Bridge dataset~\citep{ebert2021bridge} for our tasks. \textbf{Top:} Two-phase open microwave; \textbf{Bottom:} Sweep beans into pile with tool.}}
    \label{fig:complex_tasks}
    \vspace{-0.5cm}
\end{figure}

\begin{figure}
  \begin{center}
    \includegraphics[width=0.8\linewidth]{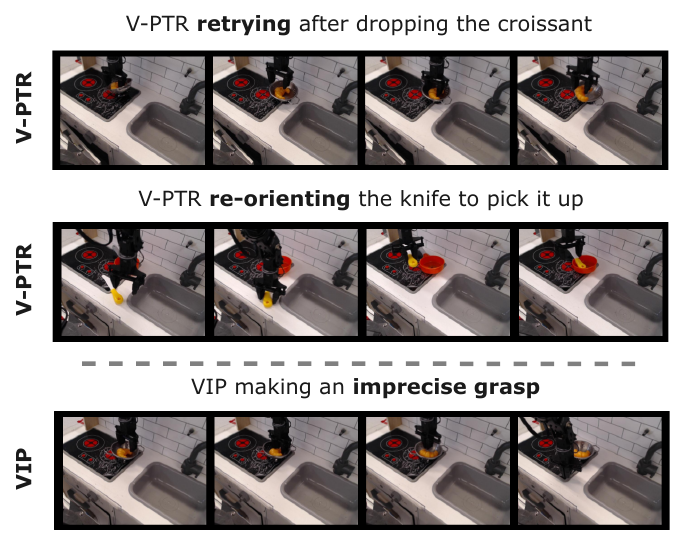}
    \vspace{-0.4cm}
  \end{center}
  \caption{\footnotesize{\textbf{Visualizing qualitative performance of \systemname\ and VIP}. Here we show rollouts for \systemname\ (top) and VIP (bottom) on the real robot manipulation tasks. \systemname\ carefully executes the task by orienting the gripper to match the object and retrying on failure whereas VIP grasp objects without this re-orientation, leading to failure.}}
  \label{fig:filmstrip}
  \vspace{-0.5cm}
\end{figure}

\textbf{Real-world setup.} We conduct our experiments on a WidowX robot platform. We perform video pre-training on Ego4D~\citep{grauman2022ego4d}, an egocentric video dataset consisting of 4M transitions of humans attempting diverse tasks in the real world, using the same pre-processed subset from prior works~\citep{nair2022r3m,ma2022vip}. Then, for multi-task robot pre-training, we utilize the subset of the Bridge dataset~\citep{ebert2021bridge,walke2023bridgedata} used by prior work~\citep{kumar2022pre}, a dataset with 150K transitions of various manipulation task demonstrations on a WidowX robot in toy kitchens. Downstream, we fine-tune on several tasks on a WidowX robot, in a \textit{previously unseen} toy-kitchen workspace. For each target task, we collect $10$ demonstrations using teleoperation, with a range of distractor objects and object variability. Solving these tasks requires skills such as picking and placing a variety of objects, using tools to accomplish tasks (e.g., sweeping), and two-phase door opening (Figures~\ref{fig:complex_tasks} and \ref{fig:filmstrip}). 

\textbf{Comparisons to prior methods.} We compare \systemname\ to approaches that do not utilize video data (PTR~\citep{kumar2022pre}, BC~\citep{ebert2021bridge}), as well as other methods for video pre-training (R3M~\citep{nair2022r3m}, MVP~\citep{xiao2022masked,radosavovic2022real}, and VIP~\citep{ma2022vip}). Following the protocols in these prior works, we fine-tune the R3M and MVP representations with imitation learning on multi-task and target robot data (phases 2 and 3), and evaluate the VIP representation both with reward-weighted imitation learning and with downstream offline RL via CQL, to provide an apples-to-apples comparison with \systemname. We evaluate three versions of VIP: \textbf{(i)} ``VIP$_{\text{frozen}}$'', which freezes the pre-trained representation learned from video data during fine-tuning, \textbf{(ii)} ``VIP'', which continues to fine-tune on robot data, and \textbf{(iii)} ``VIP$_\text{reward}$'', which not only utilizes the pre-trained representation for initializing the policy but also uses distances between representations of current and future frames as a reward shaping for downstream offline RL via reward-weighted imitation, following \citep{ma2022vip}. When using language task specification, we compare to language conditioned variants of BC and PTR. 

\begin{table*}[t]
    \vspace{0.4cm}
    \centering
    \tablestyle{3pt}{1.0}
    \resizebox{\linewidth}{!}{%
    \begin{tabular}{c | c |c || c | cccc c c}
    \toprule
    & & & &  \multicolumn{4}{c|}{\textbf{Video pre-training}} & \multicolumn{1}{c|}{\textbf{No videos}} &  \multicolumn{1}{c}{\textbf{No robot data}} \\
    \cmidrule{3-10}
      & & {\textbf{Task}} & \textbf{\systemname\ (Ours)} & \textbf{R3M+BC} & \textbf{MVP+BC}& \textbf{VIP+CQL} & \textbf{VIP$_{\text{frozen}}$+CQL} & \textbf{PTR} & \textbf{\systemname\ w/o phase 2} \\
    \shline %
    \parbox[t]{2mm}{\multirow{3}{*}{\rotatebox[origin=c]{90}{\!\!\!\!Scenario 1}}} & \multirow{3}{*}{\rotatebox{90}{\parbox{1cm}{}}} & 
    Croissant from bowl & 7 / 12 & 0 / 12 & 4 / 12 & 2 / 12 & 0 / 12 & 3 / 12 &  5 / 12   \\
    & & Sweet potato on plate & 6 / 12 & 0 / 12 & 1 / 12 & 0 / 12 & 0 / 12 & 1 / 12 & 1 / 12\\
    & & Knife in pot & 6 / 12 & 0 / 12 & 0 / 12 & 0 / 12 & 0 / 12 & 0 / 12 & 0 / 12 \\
    & & Cucumber in pot & 5 / 12 & 0 / 12 & 1 / 12 & 0 / 12 & 0 / 12 & 1 / 12 & 1 / 12  \\
    \cline{3-10}
    & & \textbf{Total} & \textcolor{red}{24 / 48}  & 0 / 48 & 6 / 48  & 2 / 48 & 0 / 48 & 5 / 48 &  7 / 48  \\
    \shline
     \parbox[t]{2mm}{\multirow{3}{*}{\rotatebox[origin=c]{90}{\!\!\!\!\!\!\!\!\!\!\!Scenario 2}}} & \multirow{3}{*}{\!\!\rotatebox{90}{\parbox{1cm}{\centering with\\ \!\!\!\!\!\!distractor~~~ \\ \!\!\!\!\!objects}}}  & Croissant from bowl & 8 / 12 & 0 / 12  & 3 / 12 & 2 / 12  & 0 / 12  & 0 / 12 & 3 / 12 \\
    & & Sweet potato on plate & 4 / 12 & 0 / 12 & 2 / 12 & 0 / 12  & 0 / 12  & 1 / 12  & 2 / 12 \\
    & & Knife in pot & 4 / 12 & 0 / 12  & 0 / 12  & 1 / 12  & 0 / 12 & 0 / 12 & 0 / 12 \\
   & & Cucumber in pot & 4 / 12 & 0 / 12 & 0 / 12  & 1 / 12 & 0 / 12  & 0 / 12 & 1 / 12  \\
   \cline{3-10}
    & & \textbf{Total} & \textcolor{red}{20 / 48}  & 0 / 48 & 5 / 48  & 4 / 48 & 0 / 48 & 1 / 48 &  6 / 48  \\
   \shline
     \parbox[t]{2mm}{\multirow{4}{*}{\rotatebox[origin=c]{90}{Scenario 3}}} & \multirow{3}{*}{\rotatebox{90}{\parbox{1cm}{\centering novel target objects}}}  & Carrot & 2 / 3 & 0 / 3 & 0 / 3 & 1 / 3 & 0 / 3 & 0 / 3 & 2 / 3  \\
    & & Cucumber & 1 / 3 & 0 / 3 & 0 / 3 & 1 / 3 & 0 / 3 & 0 / 3 & 1 / 3 \\
    & & Ice-cream & 0 / 3 & 0 / 3 & 0 / 3 & 1 / 3 & 1 / 3 & 1 / 3 & 0 / 3 \\
    \cline{3-10}
    & & \textbf{Total} & \textcolor{red}{3 / 9}  & 0 / 9   & 0 / 9  & \textcolor{red}{3 / 9} & 1 / 9 & 1 / 9 &  \textcolor{red}{3 / 9}  \\
    \shline %
    \end{tabular}}
    \caption{\footnotesize{\textbf{Task success rates of \systemname\ and prior methods} on several manipulation tasks over 12 trials (best-performing method indicated in red). Note that \systemname\ outperforms all prior methods, including those approaches that do not fine-tune the learned representation, use imitation learning for downstream control, or do not incorporate video data.}}
    \label{tab:regular_evals}
    \vspace{-0.35cm}
\end{table*}

\begin{table}[h]
    \centering
    \tablestyle{3pt}{1.0}
    \begin{tabular}{c || c | c c | c}
    \toprule
    & & & & \textbf{No CQL} \\
    \cmidrule{1-5}
      {\textbf{Task}} & \textbf{\systemname} & \textbf{VIP~\citep{ma2022vip}+CQL} & \textbf{PTR~\citep{kumar2022pre}} & \textbf{VIP$_\text{reward}$~\citep{ma2022vip}} \\
    \shline %
    Open Microwave & 5 / 12 & 2 / 12 & 0 / 12 & 0 / 12 \\
    Sweep Beans & 6 / 12 & 5 / 12 & 2 / 12 & 2 / 12 \\
    \shline %
    \end{tabular}
    \vspace{-0.1cm}
    \caption{{\footnotesize{\textbf{Performance of \systemname, VIP, and PTR on more complex tasks}. V-PTR outperforms PTR as well as VIP variants that use downstream CQL or BC weighted by the reward shaping from \citet{ma2022vip}}.}}
    \label{tab:complex_table}
    \vspace{-0.2cm}
\end{table}

\subsection{Real-World Results}
\label{sec:real_world}

We evaluate \systemname\ and comparisons in three testing scenarios. In \textbf{Scenario 1}, we evaluate the performance of the policy, varying the robot's initial pose and the position of objects in scene. In \textbf{Scenario 2}, we evaluate the policy in the presence of novel distractor objects in the scene. We note that some of the target data does contain distractor objects, but we use a different set of distractor objects during evaluation than those seen in training. Finally, in \textbf{Scenario 3}, we test the ability of the learned policy to manipulate novel target objects that were never seen during training.   

We evaluate different methods on four pick-and-place tasks in Table~\ref{tab:regular_evals}. Observe that \systemname\ outperforms all other prior approaches, and in some tasks (e.g., ``place knife in pot'') is the only method that produces any successful trials. We observed that all of these methods do learn behavior that \emph{attempt} solve the task, for example by moving toward relevant objects in the scene, but do not eventually succeed due to either imprecise localization of target objects or a prematurely executed grasp attempt. On the other hand, we found that in several scenarios \systemname\ is able to re-orient the gripper before executing a grasp (e.g., Fig.~\ref{fig:filmstrip}).

Next, we evaluate \systemname\ with distractor objects in Table~\ref{tab:regular_evals} (Scenario 2). Adding distractors reduces performance for every method (as one may expect), but \systemname\ still exhibits the smallest degradation compared to the next best approach. To study generalization to novel target objects (Scenario 3), we also consider a ``take [object] from bowl'' task, and replace the target object with unseen target objects at evaluation. Performance is low for all comparisons in Table~\ref{tab:regular_evals}, but \systemname\ succeeds $50\%$ of the time with new objects.

\begin{figure}
\centering
\includegraphics[width=0.8\linewidth]{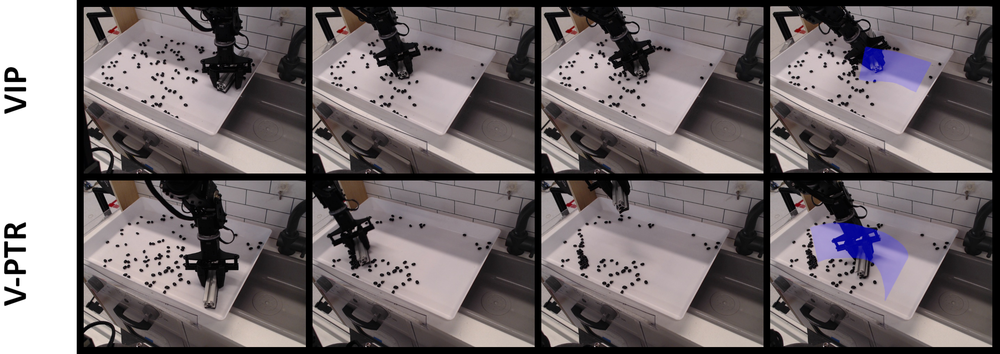}
{\caption{\label{fig:sweep_area} \footnotesize{Examples of areas swept by VIP~\citep{ma2022vip} (\textbf{top}) and \systemname\ (\textbf{bottom}) methods. \systemname\ sweeps a much larger area (blue), and consistently begins a second sweep, whereas VIP~\citep{ma2022vip} is too slow to sweep a second time.}}}
\end{figure}

\textbf{Comparisons to VIP on more complex tasks.} We compare V-PTR in more detail to VIP (which also uses similar video data) on manipulation tasks that require learning more complex skills: ``open microwave'' and ``sweep beans'' in Table~\ref{tab:complex_table}. Specifically, we compare to different variants of VIP discussed above (VIP+CQL; VIP$_\text{reward}$) and find that V-PTR outperforms these variants. Qualitatively in Figure~\ref{fig:sweep_area}, we observe that on the ``sweep beans'' task, V-PTR sweeps a larger area than the VIP policy, which is too slow to execute the sweep motion a second time. This corroborates our analysis (Figure \ref{fig:value_function} and \ref{fig:gradcam}) that value functions trained on the V-PTR representation tend to have lower error than those utilizing VIP representations. 

\textbf{Language-based task specification.} We next study how \systemname{} works when the robot pre-training data is labeled with natural language descriptions (a more general format) instead of task identifiers. 
To handle language, we first encode the task description into an embedding vector using the pre-trained language encoder from GRIF~\citep{myers2023goal}, which has been shown to be effective at learning BC policies. The Q-function and the policy then utilize these embedding vectors in lieu of the one-hot task identifiers, processing them identically as before. In Table~\ref{tab:language_table}, we compare V-PTR to imitation learning and PTR with language embeddings, and find that V-PTR improves over both of these methods by around 50\%, indicating that V-PTR can leverage downstream language.

\begin{table}[h]
    \centering
    \vspace{-0.15cm}
    \tablestyle{3pt}{1.0}
    \begin{tabular}{c c c c | c}
    \toprule
    & & & & \textbf{No language} \\
    \cmidrule{2-5}
      {\textbf{Task}} & \textbf{\systemname\ (language)} & \textbf{BC} & \textbf{PTR} & \textbf{\systemname\ (one-hot)} \\
    \shline %
    Croissant  & 7 / 12 & 3 / 12 & 5 / 12 & 7 / 12 \\
    Sweet potato  & 6 / 12 & 2 / 12 & 6 / 12 & 6 / 12 \\
    Knife in Pan & 5 / 12 & 0 / 12 & 3 / 12 & 6 / 12 \\
    Cucumber in Pot & 9 / 12 & 0 / 12 & 3 / 12 & 5 / 12 \\
    Open Microwave & 9 / 12 & 1 / 12 & 0 / 12 & 5 / 12 \\
    Sweep Beans & 8 / 12 & 2 / 12 & 4 / 12 & 6 / 12 \\
    \hline
    Total & \textcolor{red}{\textbf{44 / 72}} & 8 / 72 & 21 / 72 & 35 / 72\\
    \shline %
    \end{tabular}
    \vspace{-0.1cm}
    \caption{{\footnotesize{\textbf{Performance of language-conditioned \systemname\ compared to language-conditioned BC and PTR} for the six manipulation tasks that we study. Note that \systemname\ outperforms both of these prior methods, indicating that V-PTR can favorably learn in settings with alternate forms of task specification.
    }}}
    \label{tab:language_table}
    \vspace{-0.35cm}
\end{table}

\subsection{Visualizations and Diagnostic Experiments}

We now analyze \systemname\ more carefully by visualizing the learned features from video-pretraining, probing the generalization of the system, and assessing the quality of value estimation in downstream offline RL.

\begin{figure}[t]
\centering
\includegraphics[width=0.7\linewidth]{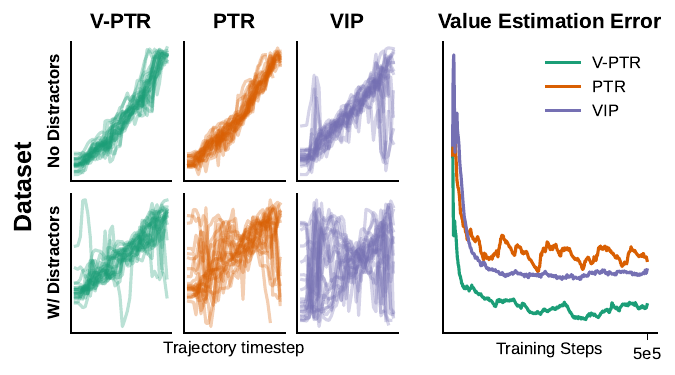}
\vspace{-0.55cm}
\caption{\footnotesize{\textbf{Visualizing the learned values $V(\bs_t)$ w.r.t. time-step $t$} on rollouts from training data (\textbf{top}), held-out data (\textbf{middle}), and rollouts with distractor objects (\textbf{bottom}) obtained after multi-task pre-training in phase 2 ($V(\bs_t)$ is computed using the average of the multi-task Q-value under actions sampled from the learned policy). Note that values trained by PTR and VIP tend to be highly non-smooth, especially on held-out rollouts with novel distractors, whereas \systemname\ produces smooth value functions.}}
\label{fig:value_function}
\vspace{-0.45cm}
\end{figure}
\textbf{Video pre-training via \systemname\ improves target value estimation.} We visualize the learned value function on frames from different rollouts in Figure~\ref{fig:value_function} (left), where the true value function should monotonically increase from initial states ($t=0$) to the final states ($t=30$) of a successful rollout. We visualize the downstream value $V(\bs_t)$ for two kinds of rollouts: \textbf{(i)} rollouts from training data and \textbf{(ii)} rollouts with distractor objects. While all three visualized methods -- PTR, VIP, and \systemname -- are able to learn smooth value functions on the data without distractor objects, in the presence of distractors, the value functions trained by PTR and VIP are non-smooth and non-monotonic (bottom). In contrast, V-PTR representations are able to induce higher-quality downstream value functions.
\begin{figure}
  \begin{center}
    \includegraphics[width=0.65\linewidth]{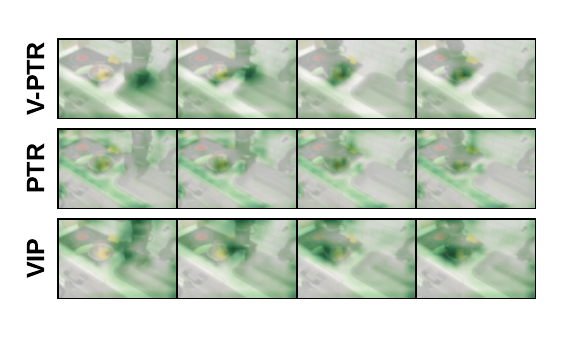}
  \end{center}
  \vspace{-0.7cm}
  \caption{\footnotesize{\textbf{Grad-CAM visuals superimposed on frames from robot data}. Regions highlighted in green denote patches of the observation with the most significant influence on the learned policy. Without PTR and VIP, background areas in the image exert a significant influence on the output of the learned policy. In contrast, initializing the policy with the video pre-trained representation obtained from V-PTR is able to focus more on gripper and object positions, crucial for solving the task.}}
  \label{fig:gradcam}
  \vspace{-0.3cm}
\end{figure}

Next, we more precisely measure the ability of \systemname\ to fit downstream value functions. We train a SARSA value function (for which we may compute a closed-form optimal solution) on top of frozen pre-trained representations from PTR (no video data), VIP~\citep{ma2022vip}, and \systemname, and report the error between the predicted value estimate and the ground-truth value on a held-out dataset in Figure~\ref{fig:value_function} (right). \systemname\ attains the smallest fitting error compared to both PTR and VIP.

\textbf{What kind of visual cues do representations trained by \systemname\ capture?} We probe which parts of the image influence the output of the learned policy for \systemname\ and other baselines, by utilizing Grad-CAM~\citep{selvaraju1610grad} to mark patches of the frame that influence the output of the learned policy in green in Figure~\ref{fig:gradcam}. We observe that \systemname\ policies discard the scene background and focus on cues relevant for robot control (e.g., object, gripper positions), while PTR and VIP place higher focuses on the scene background. This evidence provides an explanation for why \systemname\ robustly attains better performance in our experiments.

\begin{figure}[h]
  \begin{center}
    \includegraphics[width=0.75\linewidth]{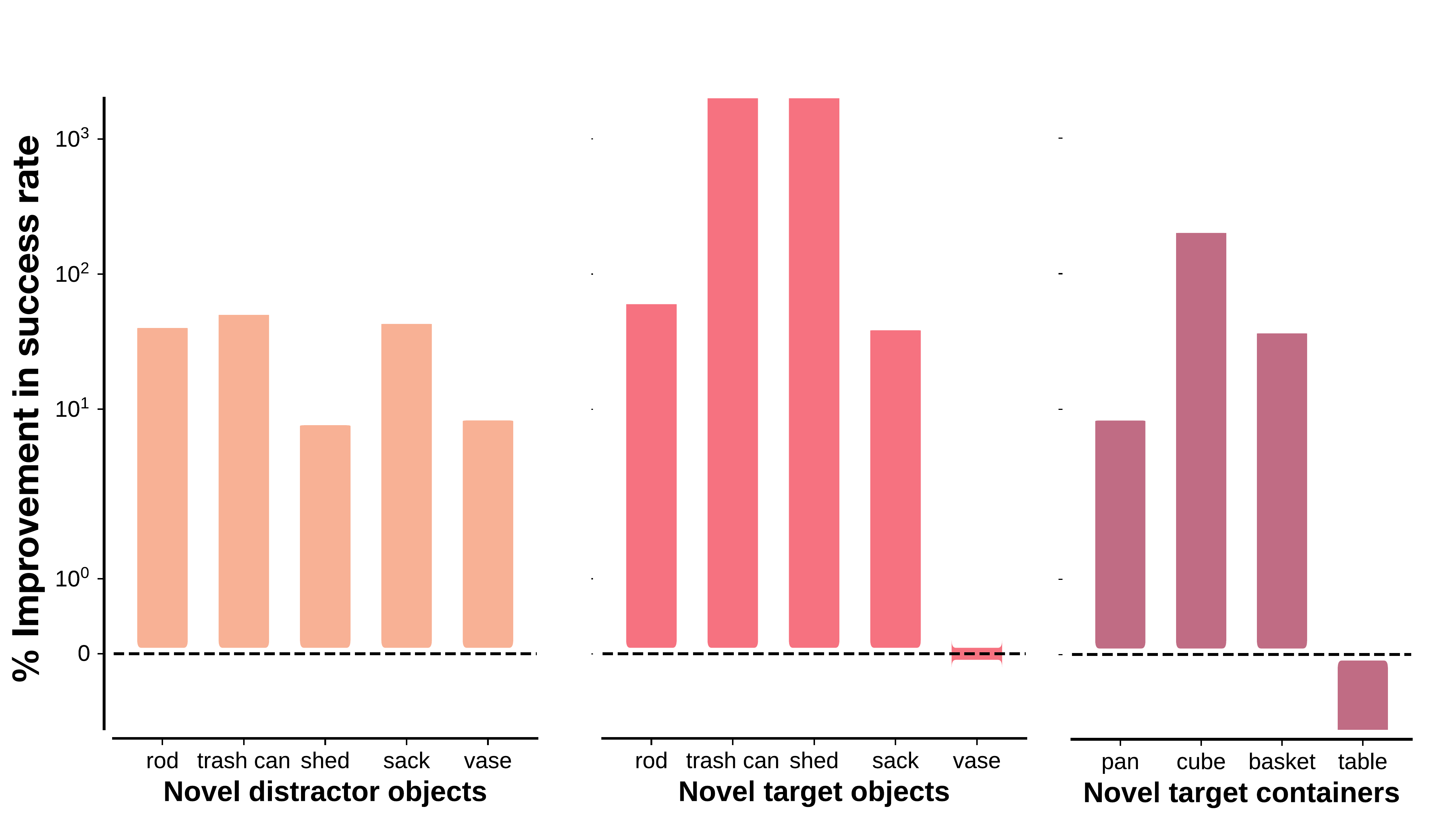}
  \end{center}
  \vspace{-0.3cm}
  \caption{\footnotesize{\textbf{Simulation results}. Percentage improvement in success rates of \systemname\ over PTR~\citep{kumar2022pre}, that does not utilize any video data for pre-training. y-axis is in log scale. In all but one scenario, V-PTR improves performance.}}
  \label{fig:simulation}
  \vspace{-0.3cm}
\end{figure}
\textbf{Simulation results.} We also evaluate \systemname\ in a simulated pick-and-place task~\citep{kumar2021workflow, kumar2022pre} where generalization to new objects can be measured more precisely. The task is to place a target object into a container, in the presence of a distractor object. Like our real-world experiments, we pre-train on video data from Ego4D, and use simulated robot demonstrations for multi-task pre-training and task fine-tuning. We evaluate \systemname\ under multiple conditions that require policies to behave robustly in the presence of distractor objects and generalize to novel target objects and workspaces. 
In Figure~\ref{fig:simulation}, we present our results in terms of the percentage improvement in success rates obtained by \systemname\ over those of  PTR~\citep{kumar2022pre} (with no video data). Consistently, across all but one scenario with a novel target container, \systemname\ improves over PTR, demonstrating that value-based video pre-training on Ego4D boosts performance and robustness of robotic RL.

\vspace{-0.1cm}
\section{Discussion and Conclusion}
\label{sec:conclusion}

In this paper, we designed a robotic system, \systemname, that uses value function pre-training on the large-scale Ego4D video dataset~\citep{grauman2022ego4d} and the robot Bridge dataset~\citep{ebert2021bridge} to improve the performance of policies learned downstream. While \systemname\ outperforms prior methods for learning from video data, we found that all the current methods remain sensitive to deviations in workspace height, camera angle, and robot configurations. There also exist many opportunities to scale, whether incorporating multi-robot datasets, larger human video datasets with language, or larger models. Nevertheless, our evaluations and diagnostic experiments indicate the promise of using RL-like value pre-training on video data for improving the general ability of robot learning algorithms.

\section*{Acknowledgements}
We thank Jason Ma, Karl Schmeckpeper, Homer Walke, Mitsuhiko Nakamoto, and Ria Doshi for their help with setting up baseline methods and for debugging the robot setup. We thank Katie Kang, Colin Li, Oleg Rybkin, Ajay Sridhar, and all other members of the RAIL lab at UC Berkeley for their feedback on an earlier version of this paper and for informative discussions. This work is supported by an NSF graduate fellowship and Office of Naval Research (N00014-21-1-2838). We also thank the TRC programs from Google cloud for providing us with TPU resources that were critical for enabling this research.

\bibliography{example}

\newpage
\appendix

\appendix

\part*{Appendices}

\section{Full System Description: \systemname}
\label{app:method_details}

In this section, we will provide further details regarding the 3 phases of \systemname\. In particular, we will describe the video pre-training with ICVF ~\citep{ghosh2023reinforcement} as well as Multi-robot pretraining with PTR~\citep{kumar2022pre}, describing both the networks present during training as well as the objective function.

\vspace{-0.2cm}
\subsection{Video Pre-Training with ICVF}

\textbf{Network.} We define the visual backbone $f_\theta(\bs)$ to be a ResNet50-v2 model, which outputs a $2048$-dimensional vector. The ICVF heads $\phi, \psi, T$ are 2-layer MLPs with hidden dimensions of $256$ and a final dimension of $256$. The ICVF is then defined as $V(\bs, \bg, \bz) = \phi(f(\bs))^\top T(f(\bz)) \psi(f(\bg))$ for video frames $\bs, \bg, \bz$.

\subsection{Multi-robot pretraining with PTR}

\textbf{Networks.} We mirror the experimental setup of \citep{kumar2022pre}, with no modifications except for a different visual backbone. PTR uses CQL \citep{kumar2020conservative} as the base offline RL method, meaning it trains two Q functions, a separate policy, and delayed target copies of the Q-functions. Each network is represented by a 3-layer MLP head with width of 256, after the visual representation. To make comparisons between PTR and V-PTR exact, we also use separate encoders for the actor and critic, although both are initialized at the beginning of Phase 2 using the same visual representation as learned during Phase 1. We refer to \citep{kumar2022pre} for a more complete description.

\section{Environment Setup and Dataset Details}
\label{app:env_setup}
In this section, we will describe the setup for our experimental results with respect to both the real-world experiments and the sim diagnostic experiments. There will be a description of the task setup as well as the corresponding datasets associated with the tasks.

\subsection{Description of State and Action Space}
{Our state and action description follows that of PTR~\citep{kumar2022pre}. The state is a 128$\times$128 RGB image captured from an over-the-shoulder camera, a one-hot vector identifying which task is to be performed, the position and rotation of the end-effector, and how much the gripper is closed. The action is the translational and angular velocity of the robot end-effector, as well as how much the gripper is closed. Rotations and angular velocities are expressed using Euler angles.} 

\subsection{Description of Real Robotic Setup}
\begin{figure}[h]
    \centering
    \includegraphics[width = \textwidth]{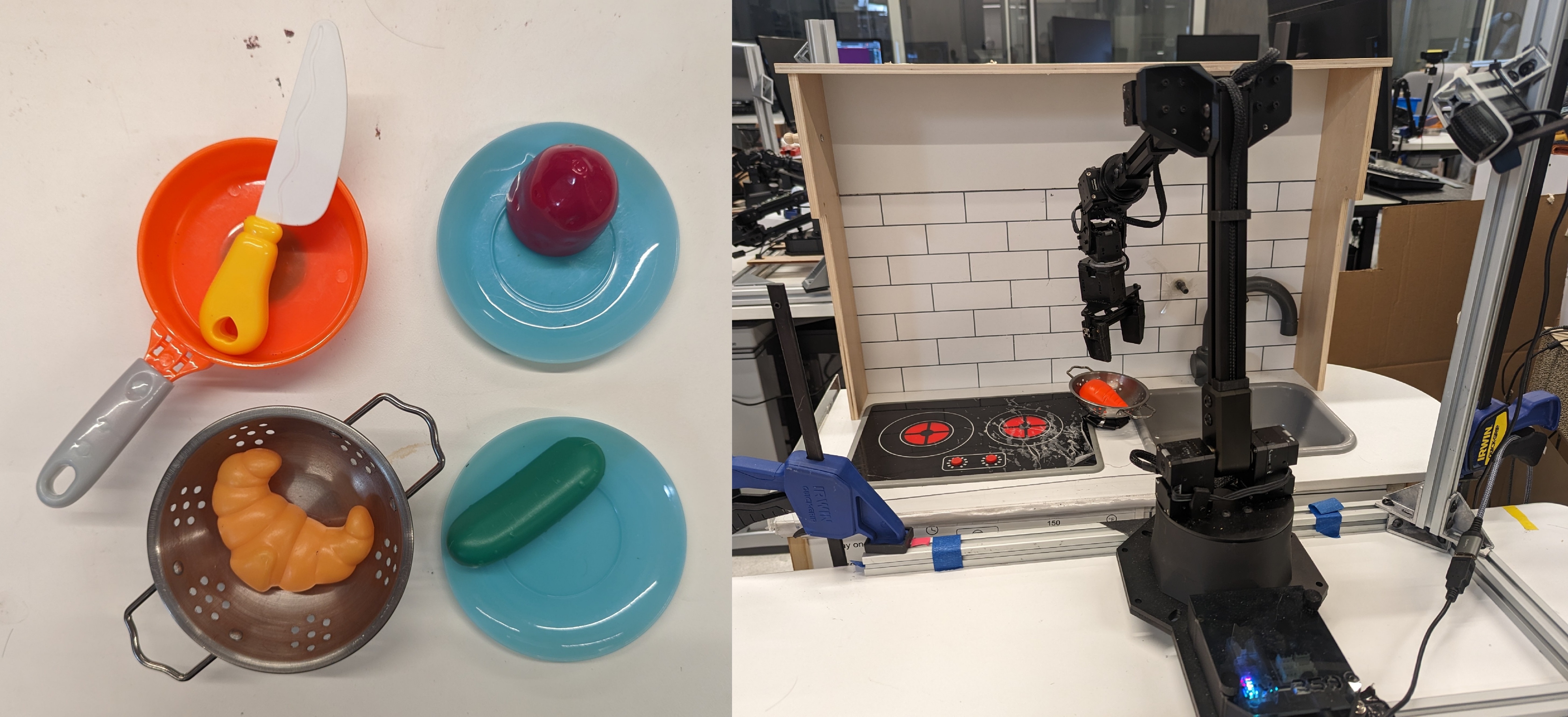}   
    \caption{\small \textbf{Real-robot experiments}. We utilize the setup from the Bridge dataset~\citep{ebert2021bridge}. The bridge dataset is collected on a 6-DoF WidowX 250 robot, with a fixed side camera placed in diverse toy-kitchens. Observations for the tasks consist of one $128\times 128$ RGB image from a side camera, as well as robot proprioception. \textbf{Left:} task objects and containers for the tasks that we study. \textbf{Right:} Evaluation setup pick place environment.}
    \label{fig:ptr_setup}
\end{figure}

We mirror our real-world experimental setup from Bridge~\citep{ebert2021bridge} and PTR~\citep{kumar2022pre}. In particular, we designed and evaluated the performance of our method under several distinct conditions in 2 different toy kitchen domains. The robot that is used is a 6-DoF WidowX 250 robot with a fixed side camera. The scene in which the robot was placed in was a toy kitchen with elements such as stove tops, sinks, microwaves, and food objects found in the scene. A picture of our setup is found in Figure~\ref{fig:ptr_setup}. 

We evaluate several pick place tasks, in which the agent has to place an object into a container amongst distractors found in the scene. Distractor objects in this scene can include other objects and containers that may have even been shown to be corresponding to different tasks. 

\subsection{Description of our Simulation Task}

\begin{figure}[t]
  \begin{center}
    \includegraphics[width=0.29\textwidth]{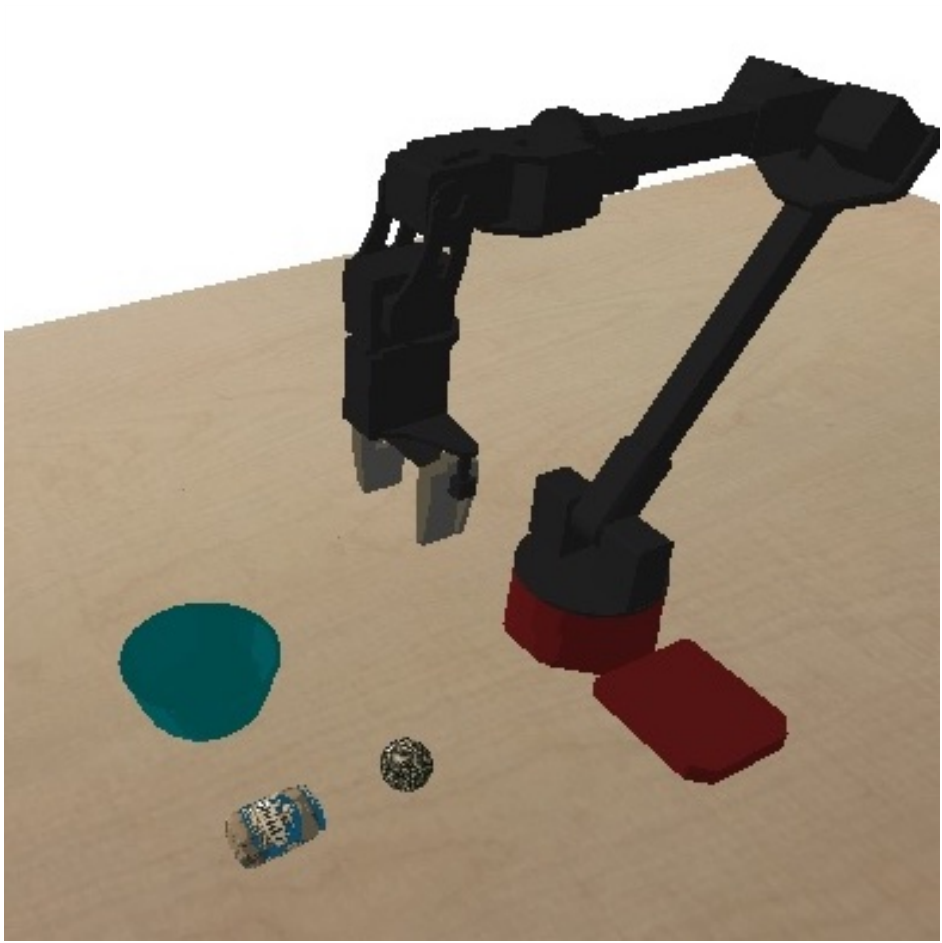}
  \end{center}
  \caption{\footnotesize{\textbf{Observations for the simulated pick-and-place} tasks consist of one $128\times 128$ RGB image from a top-down camera, as well as a proprioceptive robot state.}}
  \label{fig:cog_obs}
  \vspace{-0.2cm}
\end{figure}
       
Our manipulation tasks in simulation are modified versions of the pick-and-place tasks from \citet{singh2020cog}, which were then extended by \citet{kumar2022pre}. This environment requires controlling a simulated WidowX 250 arm which is placed in front of a single bin with an object to place in the bin and a distractor. These objects are selected from a pool of 10 objects, with no overlap between the objects to be placed in the bin and the distractors. The goal of the task is to pick up the object and place it in the bin successfully. The setup is designed in the PyBullet simulation framework which models the physics of the robotic arm, the objects in the scene, as well as the interactions between them. A picture of our setup is found in Figure~\ref{fig:cog_obs}. The dataset is collected with a scripted policy that reaches for the object of interest, grasps it with some likelihood, and places it in the bin. This collected dataset had an overall success rate of 30\%.

\subsection{Description of Video Pre-Training Dataset}
The Video Pre-training Dataset that we utilize in Stage 1 of our method, \methodname, is Ego4D~\citep{grauman2022ego4d}. We used the same pre-processed Ego4D dataset as in R3M~\citep{Nair2022R3MAU} and VIP~\citep{ma2022vip}. In particular, long videos that are raw in nature are clipped to be shorter consisting of 10-150 frames. From here, the clipped video is decomposed into individual frames that are pre-processed with a random crop at the video level. The frame is then resized and center-cropped to be of size $224 \times 224 \times 3$. These processed video frames are then fed into the replay buffer to be individual transitions for the ICVF objective~\citep{ghosh2023reinforcement}.

\subsection{Description of Real-World Multi-Task Robot Datasets}
For the pre-training and target datasets in Stage 2 of \systemname\ for the real-world multi-task training, we utilize subsets of the Bridge Dataset~\citep{ebert2021bridge}. Mirroring the setup of Scenario 3 in PTR~\citep{kumar2022pre}, the pre-training data comprises of all pick-and-place data found in the bridge dataset except for any demonstration data collected in the toy kitchen that was evaluated on. For the target dataset, we collect 44 new demonstrations for each of the 4 tasks: Removing Croissant from Colander, Placing Knife in Pot, Placing Sweet Potato on Plate, and Placing Cucumber in Bowl. A total of 218 successful demonstrations were collected with a human demonstrator using an Oculus Quest Headset for Phase 3 of \systemname.

\section{Evaluation Details }
In this section, we will provide how real and simulated environments were evaluated fairly across our method and baselines. This protocol is similar in nature to the one presented in PTR~\citep{kumar2022pre}.

\label{app:eval_details}
\subsection{Evaluation Protocol for Real-World Experiments}
To evaluate a policy in the real world, we loop through 4 starting gripper transformations that move the gripper to the left front, right front, left rear, and right rear of the kitchen environment. For each of these starting transformations, we evaluate a total of 3 times: once with the object to pick up directly underneath the gripper and twice with it shifted slightly so that the policy cannot simply grip from the starting location. For each of these evaluations, we let the policy run for 60 seconds. 

For our experiments testing generalization to novel distractor objects, we place two novel distractor objects into the scene. We shift the locations of these objects between each evaluation and switch out the objects themselves when we change start transforms so that each combination of distractors is evaluated 3 times. 

For new target object experiments, which we only conducted for the croissant out of colander task, we replace the croissant with a new object never seen at training time. We switch out the target object for each start transform so that each new target object is evaluated 3 times. 

When taking objects out of containers, we do not count knocking over the container so that the object falls out as a success - the gripper must lift up the object in the air and set it outside the container. 

\subsection{Evaluation Protocol for Simulation Experiments}

The multi-task robot dataset in these experiments consists of 1500 pick-place trajectories for 6 target objects. We created three small target datasets to test generalization along different axes: \textbf{(a)} diverse target objects to pick and place, \textbf{(b)} diverse containers holding the target objects, and \textbf{(c)} diverse distractor objects in the scene. Each of these datasets contains 10 successful demonstrations of the target task, spread across 5 different objects. We assign a unique one-hot task id for each of the source target objects and a single task id for the target dataset. Then, we train our system, \systemname, and a baseline PTR policy using multi-task CQL with hyperparameters $\alpha=5$ and target mixing ratio $\tau=0.7$.

When evaluating zero-shot generalization, we test each of these policies along the axis of generalization present in their target datasets. In particular, for the policies trained with diverse targets, diverse distractors, and diverse containers, we test generalization to 5 new targets, distractors, and containers respectively, which are unseen in the source and target datasets. Each evaluation is performed with 20 rollouts and an environment time limit of 40.

\section{Experimental Details for \systemname~and Baseline Methods}

\begin{table}[t]
\centering
\scalebox{0.99}{
\begin{tabular}{p{3.cm} p{3.cm} p{3.cm}  p{3.cm}  p{3cm}}
  \toprule
    \textbf{Hyperparameters} & \textbf{Simulation}  & \textbf{Pick Place}\\         
  \midrule
    $\alpha$ & 0.1, 1,5,10  & 0.1, 1,5,10\\
   policy architecture &  ResNet-50, ViT-B &  ResNet-50, ViT-B\\
   critic architecture &  ResNet-50, ViT-B &  ResNet-50, ViT-B\\
   policy learning rate &  1e-4  & 1e-4\\
   critic learning rate &  3e-4  & 3e-4\\
    reward scale &  11 & 11\\
    reward bias &  -1  & -1\\
    batch size &  64 & 256\\
  \bottomrule
\end{tabular} }
\vspace{0.05cm}
\caption{\footnotesize \textbf{The hyperparameters used by \systemname}. After pre-training on multi-task robot data, in the second pre-training phase \systemname\ fine-tunes the representation using multi-task CQL~\citep{kumar2020conservative} on diverse robot data. Finally the third fine-tuning phase aims to customize this policy for the target task. The above table presents hyperparameters for \systemname\ that we utilize in our experiments.}

\label{table:hyperparam_all_cql} 

\end{table}

\subsection{CQL Finetuning~\citep{kumar2022pre}}
For our second pre-training phase on multi-task robot data and fine-tuning on target data, we utilized CQL~\citep{kumar2020conservative} as the downstream offline RL algorithm. Following the design decisions in the official implementation of PTR~\citep{kumar2022pre}, we utilized a variant of Bellman backup that computes the target value by performing a maximization over target values computed for $n=4$ actions sampled from the policy at the next state (max\_q\_backup). In each domain, we swept over the alpha values of $\alpha=0.1, 1, 5, 10$. We built our code upon the CQL implementation from \url{https://github.com/Asap7772/PTR}~\cite{kumar2022pre}. 

\subsection{MVP~\citep{Xiao2022MaskedVP,radosavovic2022real}}
For masked autoencoding (MAE)-based methods such as MVP~\citep{radosavovic2022real}, we loaded in the pre-trained PyTorch Checkpoints for the ViT-Base model. For this method, we kept the pre-trained representation frozen and finetuned a Multi-Layer Perceptron (MLP) that takes in the class token of the ViT as input to the model and outputs an action dimensional vector. The output of the MAE is normalized with layer normalization. Other hyperparameters follow the CQL fine-tuning section.

\subsection{VIP~\citep{ma2022vip} and R3M~\citep{Nair2022R3MAU}}
Prior methods have shown that TD-based updates with networks with batch normalization have instabilities during training. Given this, methods such as VIP and R3M that utilize a standard ResNet-50~\citep{resnet} backbone with batch normalization do not finetune stably for methods such as CQL. For this reason, any comparisons using standard ResNet 50 either use frozen batch statistics, or are trained with a behavioral cloning objective.

All hyperparameters described above are summarized in Table~\ref{table:hyperparam_all_cql}.

\section{Reducing the Number of Fine-Tuning Demonstrations}
{We examine how reducing the number of demonstrations from 44 to 10 degrades performance in Table ~\ref{tab:10demo_evals}. \systemname\ continues to significantly outperform baselines.}

\begin{table*}[h]
    \centering
    \tablestyle{3pt}{1.0}
    \begin{tabular}{c c | c c}
    \toprule
    & &  \multicolumn{1}{c|}{\textbf{Video pre-training}} & \multicolumn{1}{c|}{\textbf{No videos}} \\
    \cmidrule{2-4}
      {\textbf{Task}} & \textbf{\systemname\ (Ours)} & \textbf{VIP~\citep{ma2022vip}+CQL} & \textbf{PTR~\citep{kumar2022pre}} \\
    \shline %
    Croissant out of Colander & 4 / 12 & 1 / 12  & 0 / 12 \\
    Sweet Potato on Plate & 5 / 12 & 0 / 12  & 0 / 12 \\
    Knife in Pan & 2 / 12 & 0 / 12  & 0 / 12 \\
    Cucumber in Pot & 4 / 12 & 0 / 12  & 0 / 12 \\
    Open Microwave & 8 / 12 & 2 / 12  & 4 / 12 \\
    Sweep Beans & 2 / 12 & 0 / 12  & 5 / 12 \\
    \shline %
    \end{tabular}
    \vspace{-0.2cm}
    \caption{{\footnotesize{Task success rates of \systemname\ and prior methods with only 10 demonstrations. As should be expected, the performances of all approaches, including ours, degrade with less data, but \systemname\ performs significantly better than other pre-training methods.}}}
    \label{tab:10demo_evals}
    \vspace{-0.35cm}
\end{table*}

\end{document}